\def\year{2023}\relax
\documentclass[letterpaper]{article} 
\usepackage{aaai22}  
\usepackage{times}  
\usepackage{helvet}  
\usepackage{courier}  
\usepackage[hyphens]{url}  
\usepackage{graphicx} 
\usepackage{natbib}  
\usepackage{caption} 
\DeclareCaptionStyle{ruled}{labelfont=normalfont,labelsep=colon,strut=off} 
\usepackage{amsmath,amssymb,amsfonts}
\usepackage{graphicx}
\usepackage{textcomp}
\usepackage{algorithm, algpseudocode}
\usepackage{subfigure}
\usepackage{threeparttable}

\usepackage{newfloat}
\usepackage{listings}
\floatstyle{ruled}
\newfloat{listing}{tb}{lst}{}
\floatname{listing}{Listing}
    \def\correspond{%
	\ifnum\value{eqfn}=0%
	\footnote{Corresponding author: Yongbing Zhang.}%
	\setcounter{eqfn}{\value{footnote}}%
	\else%
	\footnotemark[\value{eqfn}]%
	\fi%
}%

\title{HVTSurv: Hierarchical Vision Transformer for Patient-Level Survival Prediction from Whole Slide Image}
\author{
     Zhuchen Shao$^{1}$,
     Yang Chen$^{1}$,
     Hao Bian$^{1}$,
     Jian Zhang$^{2}$,
     Guojun Liu$^{3}$,
     Yongbing Zhang$^{4}$\correspond
}
\affiliations{
    \textsuperscript{\rm 1}Tsinghua Shenzhen International Graduate School, Tsinghua University\\
    \textsuperscript{\rm 2}School of Electronic and Computer Engineering, Peking University\\
        \textsuperscript{\rm 3}Computer Science and Technology, Harbin Institute of Technology\\
     \textsuperscript{\rm 4}Computer Science and Technology, Harbin Institute of Technology (Shenzhen) \\
        shaozc0412@gmail.com,  ybzhang08@hit.edu.cn
}

\usepackage{bibentry}

\begin{document}

\maketitle

\begin{abstract}
Survival prediction based on whole slide images (WSIs) is a challenging task for patient-level multiple instance learning (MIL). Due to the vast amount of data for a patient (one or multiple gigapixels WSIs) and the irregularly shaped property of WSI, it is difficult to fully explore spatial, contextual, and hierarchical interaction in the patient-level bag. Many studies adopt random sampling pre-processing strategy and WSI-level aggregation models, which inevitably lose critical prognostic information in the patient-level bag. In this work, we propose a hierarchical vision Transformer framework named HVTSurv, which can encode the local-level relative spatial information, strengthen WSI-level context-aware communication, and establish patient-level hierarchical interaction. Firstly, we design a feature pre-processing strategy, including feature rearrangement and random window masking. Then, we devise three layers to progressively obtain patient-level representation, including a local-level interaction layer adopting Manhattan distance, a WSI-level interaction layer employing spatial shuffle, and a patient-level interaction layer using attention pooling. Moreover, the design of hierarchical network helps the model become more computationally efficient. Finally, we validate HVTSurv with 3,104 patients and 3,752 WSIs across 6 cancer types from The Cancer Genome Atlas (TCGA). The average C-Index is  2.50-11.30\% higher than all the prior weakly supervised methods over 6 TCGA datasets. Ablation study and attention visualization further verify the superiority of the proposed HVTSurv. Implementation is available at: https://github.com/szc19990412/HVTSurv.
\end{abstract}

\section{Introduction}
In computational pathology, survival prediction based on gigapixels whole slide images (WSIs) is a weakly supervised learning (WSL) task involving local-level tumor microenvironment interactions  \cite{chen2021multimodal}, WSI-level tumor-related tissue interactions  \cite{abbet2020divide} and patient-level heterogeneous tumor interactions \cite{carmichael2022incorporating}. 
Multiple instance learning (MIL) is usually adopted to tackle such a WSL problem \cite{shao2021transmil,shao2021weakly}. However, bag-based representation learning in MIL still remains an open and challenging problem.\par

\begin{figure}[t]
\centering
\includegraphics[width=1.0\linewidth]{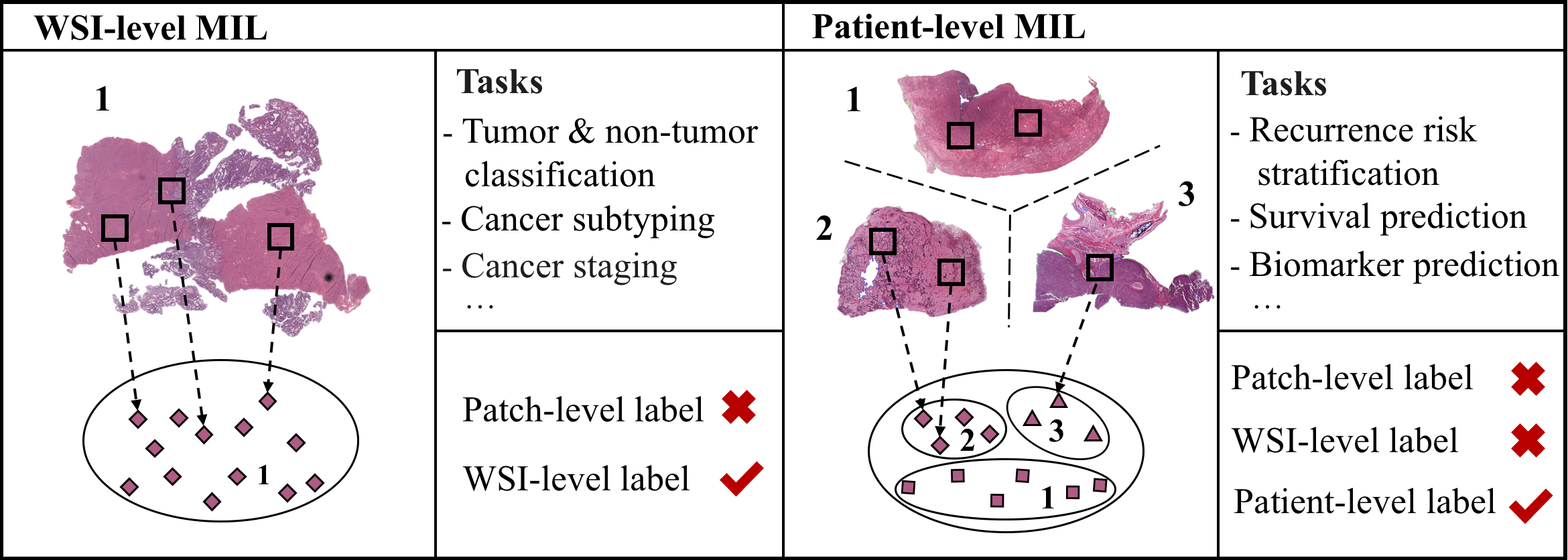}
\caption{Application of MIL in WSIs-based tasks. There is only one WSI-level bag for the WSI-level MIL, where the WSI-level label is known, and the patch-level label is unknown. For the patient-level MIL, there are one or more than one WSI-level bags where the patient-level label is known, but neither the WSI-level nor patch-level label is unknown.}
\label{figMIL}
\end{figure}

Different from natural images, WSIs have the property of high-resolution and wide field of view  \cite{srinidhi2021deep}, so the aggregation of bag-level representation will impose a great demand on computational complexity.
In addition, 
different from the WSI-level MIL problem, as shown in Fig.~\ref{figMIL}, survival prediction based on WSIs is a patient-level MIL problem \cite{fan2021learning}. 
Since the tumor may have a composite tissue structure, multiple WSIs are usually collected for patient diagnosis. Therefore, in survival prediction, we have to face two dilemmas: 1) 
multiple WSIs inevitably lead to linearly multiplied data volume; 2)~the aggregation of multiple WSI-level bags for a patient.\par
For the aggregation of the WSI-level bag, the risk information in survival prediction is often reflected in a series of histological patterns corresponding to disease progression. For example, the local-level co-localization of tumors and lymphocytes \cite{shaban2019novel} and the WSI-level metastatic distribution of sentinel lymph nodes \cite{kim2020challenge} have been shown to correlate with prognosis.
Therefore, the spatial and  contextual information must be fully considered in a WSI-level bag. Moreover, for the patient-level bag, intratumoral heterogeneity will inevitably lead to diverse tumor microenvironments in different WSIs \cite{vitale2021intratumoral}. So the patient-level contextual information between instances in different WSI-level bags must be considered, and these three-level interactions further constitute hierarchical information in a patient-level bag.

To address the challenges mentioned above, numerous works are proposed from two majority aspects: 1) computational cost;
2) spatial, contextual and hierarchical information encoding.
For the computational cost problem, random-sample-based methods \cite{huang2021integration,di2022big} and cluster-based methods \cite{yao2020whole,muhammad2021epic,shao2021weakly} are widely used.
By randomly selecting from different clusters, many cluster-based methods try to include various richer tissue types. 
However, a large number of tissue patches are discarded and usually lack structural information.
For the spatial information encoding, 
some GNN-based methods \cite{wang2021hierarchical,li2018graph,chen2021whole} adopt the topological structure to encode neighbor node information in WSI. 
In addition, Transformer-based methods \cite{huang2021integration,shao2021transmil} adopt trade-off strategies like using the sin-cos embedding  or applying convolution to implicitly encode location information. However, it is still an unsolved problem to efficiently encode 2D spatial information over such high resolution and irregularly shaped WSIs. 
For the contextual and hierarchical information encoding, patch-based graph convolutional network \cite{chen2021whole}, Nystrom self-attention \cite{shao2021transmil} and non-local attention \cite{li2020dual} are used to encode WSI-level interactions. 
There are also other methods \cite{di2022big,fan2021learning} hierarchically processing WSI and patient-level bags to encode hierarchical information.
However, limited by the vast amount of data for patient-level survival prediction, randomly sampled patches are always used in the methods above, which inevitably lose potential risk information.

In this work, we propose a hierarchical vision Transformer for patient-level survival prediction (HVTSurv) that progressively explores local-level spatial, WSI-level contextual and patient-level hierarchical interactions in the patient-level bag. The main contributions are as follows:

1) To alleviate high computational complexity,
we devise a local-to-global hierarchically processing framework. Specifically, we leverage the window attention mechanism to design the local and shuffle window block, which can significantly reduce the cost of Transformer.
Therefore, we can take advantage of all the patches in the patient-level bag.

2) We propose a new feature generation method for spatial information embedding, which can fully reflect the local characteristics in both horizontal and vertical directions. In addition, we adopt Manhattan distance in the local window block to measure the local relative position.

3) For the contextual and hierarchical information encoding, we design the local-level, WSI-level and patient-level interaction layers to hierarchically deal with survival prediction. 
Besides, we adopt a random window masking strategy in feature generation to further exploit the advantages of our hierarchical
processing framework.

4) Our HVTSurv significantly outperforms state-of-the-art (SOTA) methods over 6 public cancer datasets from The Cancer Genome Atlas (TCGA) with less GPU Memory Costs. Besides, in the patient stratification experiment, there is a statistically significant difference (P-Value $<$ 0.05) over all of the 6 cancer types. Attention visualization further verifies our conclusion from experiments.

\section{Related Work}
\subsection{Application of MIL in WSIs}
The application of MIL in WSIs can be divided into two categories. As shown in Fig. \ref{figMIL}, MIL tasks in WSIs include WSI and patient-level MIL. 
The WSI-level MIL is suitable for some tasks such as tumor\&non-tumor classification. Common solutions include instance-based methods \cite{campanella2019clinical,Xu2019CAMELAW,kanavati2020weakly,lerousseau2020weakly,Chikontwe2020MultipleIL} and embedding-based methods \cite{tomita2019attention,hashimoto2020multi,naik2020deep,lu2021data, hou2022h2}. \par 
The patient-level MIL is suitable for tasks like survival prediction that only has patient-level labels. 
Common solutions include simultaneous processing \cite{zhu2017wsisa,yao2020whole,muhammad2021epic,shao2021weakly,abbet2020divide,wang2021hierarchical,li2018graph,huang2021integration} and hierarchical processing \cite{chen2021whole,di2022big,fan2021learning} of  all the WSIs from a patient. 
Simultaneous processing methods generally consider all the WSI-level bags in a patient as one bag. In contrast, the hierarchical processing methods first aggregate features in the WSI-level bag and then further aggregate different WSI-level bags in the patient-level bag. In general, hierarchical processing methods have the potential to more effectively model the spatial information in the WSI-level bag and contextual and hierarchical information in the patient-level bag.

\begin{figure*}[ht]
\centering
\includegraphics[width=1.0\linewidth]{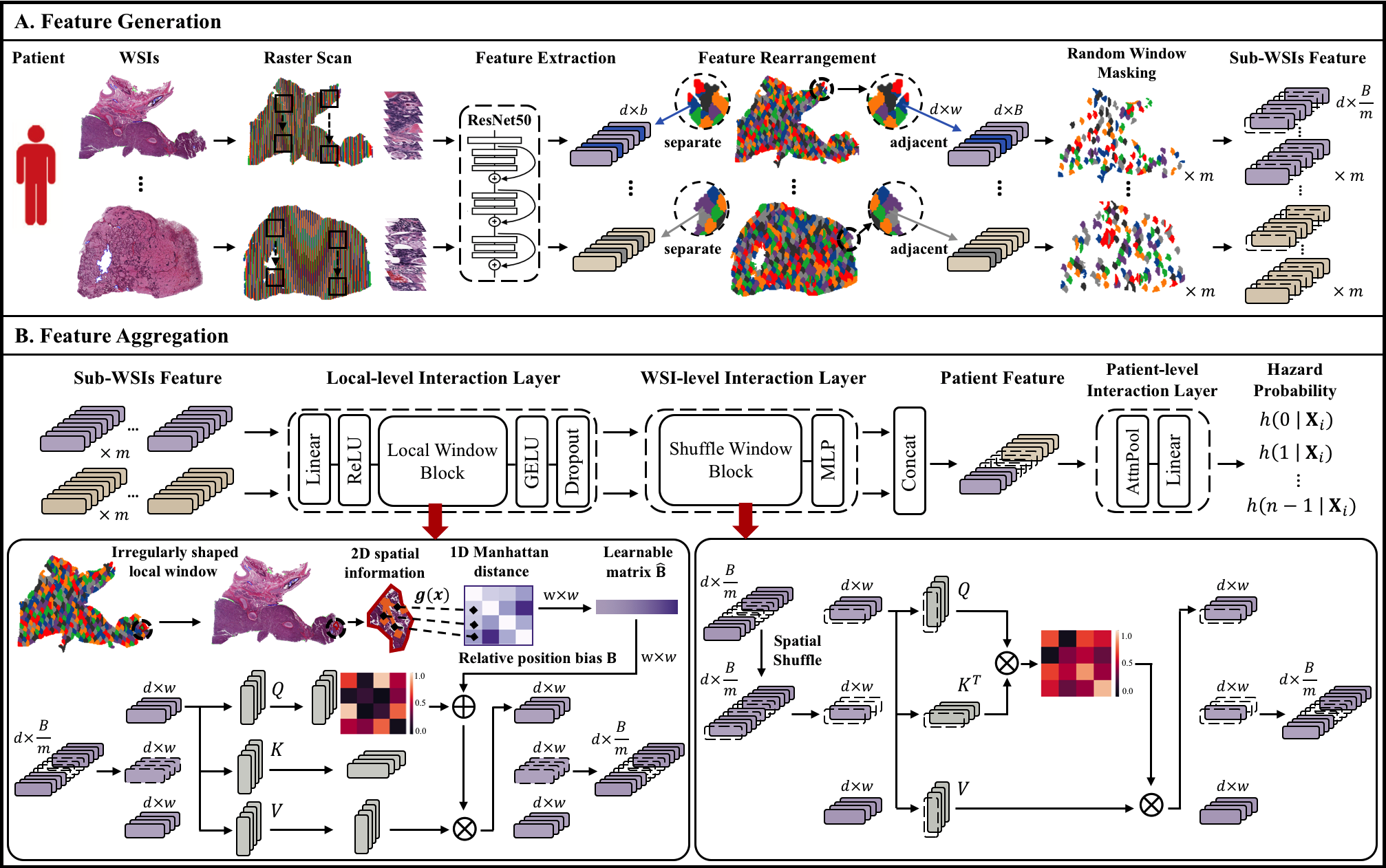}
\caption{Overview of the proposed pipeline. For all WSIs in a patient, we first segment and slice all the tissue patches and use an ImageNet pre-trained ResNet50 to extract each patch as a 1024-dimensional feature vector. Then we adopt the feature rearrangement to ensure that, after window partition, patch features in the same window are adjacent in the rearranged feature sequence. Besides, we apply a random window masking strategy to split a WSI bag into $m$ sub-WSI bags to increase the robustness of the model for tumor heterogeneity. Next, we use the generated features to perform the aggregation. For each WSI, local-level interaction layer will first encode local spatial information. And then, spatial shuffle is applied to make the model carry out similarity computation for features in different local windows. Finally, all the WSI-level features will be concatenated to perform the attention pooling, and we use the patient-level representation to predict the patient's hazard risk. The bottom is the overview of the local window block and shuffle window block. In the local window block, we add relative position bias to the self-attention calculation, and the distance is measured using the Manhattan distance. In the shuffle window block, we shuffle the patient feature in the different local windows and then perform the window partition and self-attention calculation.}
\label{figA1}
\end{figure*}

\subsection{Application of WSIs for Survival Prediction}
For the application of WSIs in survival prediction,
a two-stage framework is widely used to predict patient hazard scores: 1) sampling and encoding patches; 2) patch features aggregation.
In the first stage, constrained by limited computing resources, clustering and random sampling methods \cite{zhu2017wsisa,yao2020whole,muhammad2021epic,shao2021weakly} are widely used to select representative tissue phenotypes in WSI. 
However, these randomly sampled patches are not context-aware and lost the interactions between cells and tissue types, which are prognostic for patient survival prediction. \par 

In the second stage, many CNN based,
GNN based
and  Transformer based
methods are used.
Both Yao et al.~ \cite{yao2020whole} and Shao et al. \cite{shao2021weakly} use a small CNN network to aggregate the sampled feature.
However, CNN-based methods have inherent limitations in modeling global topological information.
For the GNN-based method, 
Chen et~ al. \cite{chen2021whole} formulate WSIs as a graph-based data structure 
to obtain hierarchical representations. Wang et al.  \cite{wang2021hierarchical} emphasize the tumor microenvironment graph construction. Di et al. \cite{di2022big} propose a big-hypergraph factorization neural network to obtain the high-order representations. 
However, the network depth limitation brought by a large amount of data makes GNN more challenging to encode WSI-level information.
For the Transformer based method, Huang et al. \cite{huang2021integration} adopt 2D sin-cos position encoding and Transformer encoder blocks to obtain the bag-level feature. There is still room for improvement in position encoding and feature aggregation to this method.

\section{Method}
Considering a set of $N$ patients $\mathbf{X}_{i}$, for $i=1, \ldots, N$, each patient $\mathbf{X}_{i}$ has one or multiple WSIs, we have follow-up label $\left(T_i, C_i\right)$, where $T_i$ stands for observation time and $C_i$ stands for survival status. The binary status $C_i \in \{0, 1\}$ indicates whether $T_i$ is a survival time ($C_i=0$) or a right-censored time ($C_i=1$). 
Our task is to predict the survival  probability based on all the WSIs
for each patient. The accuracy is measured by the consistency between the sorted survival  probability and sorted follow-up label sets. \par
To better perform survival prediction, the discrete-time survival model \cite{vale2021long,chen2021pan} is used in this paper. Briefly, we subdivide the survival time scale into $n$ intervals: $\left[t_0, t_1\right), \dots, \left[t_{n-1}, t_n\right)$, where $t_1, \dots, t_{n-1}$ define the evenly divided points of survival times for uncensored patients and ${t_0}=0, {t_n}=\infty$. 
Each patient observation time will be attributed to an interval as:
\begin{equation}
T_i=k \text { iff } T_i \in [t_k, t_{k+1}).
\label{eq1}
\end{equation}
Therefore, for each patient, the conditional hazard probability $h(k \mid \mathbf{X}_{i})$ can be defined as its failure probability in interval $[t_k, t_{k+1})$:
\begin{equation}
h(k \mid  \mathbf{X}_{i}) = P(T_i=k \mid T_i \geq k, \mathbf{X}_{i}).
\label{eq2}
\end{equation}
Survival probability $S(k \mid \mathbf{X}_{i})$ can be defined as its observation probability at least to the end of
interval $[t_k, t_{k+1})$:
\begin{equation}
S(k \mid  \mathbf{X}_{i}) = P(T_i>k \mid  \mathbf{X}_{i}) = \prod_{s=1}^{k}(1-h(s \mid \mathbf{X}_{i})).
\label{eq3}
\end{equation}

Since each patient label is known, while neither WSIs label nor patches label is unknown, survival prediction is a WSL problem, which can be solved by the MIL methods. To better predict $h(k \mid  \mathbf{X}_{i})$ from the patient-level bag, as shown in Fig. \ref{figA1}, we propose a Transformer-based framework which is composed of feature generation and feature aggregation.

\subsection{Feature Generation}
We first convert patches to features in WSI processing. Then, we adopt feature rearrangement to maintain the local 2D relative position in rearranged features. Finally, we employ the random window masking to further strengthen the contextual and hierarchical interactions in feature aggregation.
\subsubsection{WSI processing}
WSIs have gigapixels and often contain many blank regions.
We follow the CLAM~ \cite{lu2021data} processing steps to 
remove the background regions, and then cut out 256$\times$256 images at 20$\times$ resolution (0.5 $\mu$m/pixel). 
A ResNet50 model pre-trained on ImageNet is employed to embed each patch in a 1024-dimensional feature vector.

\subsubsection{Feature rearrangement}
In patient-level survival prediction, a patient may correspond to multiple WSIs.
To relieve computational cost, we employ the window attention mechanism \cite{liu2021Swin}.
Limited by irregularly shaped property of WSI, previous raster scanning method will inevitably lose correct 2D spatial information. To better reflect the local characteristics in both horizontal and vertical directions within a window, a feature rearrangement method is proposed to ensure the closeness in both directions of the 2D space after the window partition. The specific implementation is shown in Algorithm \ref{alg1}, and the Euclidean distance is used in the HNSW \cite{malkov2018efficient}.
We also present qualitative and quantitative experimental results in Appendix Fig. 1 and Appendix Fig. 2, respectively.

\begin{algorithm}
\caption{Feature rearrangement}\label{alg1}
\begin{algorithmic}
\renewcommand{\algorithmicrequire}{ \textbf{Input}}
\Require A WSI-level bag $\mathbf{H}_i=\left\{{\boldsymbol{h}}_{i,1}, \ldots, {\boldsymbol{h}}_{i,b}\right\}$, where ${\boldsymbol{h}}_{i,j} \in \mathbb{R}^{d }$ is the embedding of the $j{\text{th}}$ instance, $\mathbf{H}_i \in \mathbb{R}^{d \times b}$. Corresponding coordinates $\mathbf{Z}_i=\left\{{\boldsymbol{z}}_{i,1}, \ldots, {\boldsymbol{z}}_{i,b}\right\}$, where ${\boldsymbol{z}}_{i,j} \in \mathbb{R}^{2 }$ is the original coordinate of the $j{\text{th}}$ instance in WSI, $\mathbf{Z}_i \in \mathbb{R}^{2 \times b}$. Window size $w$.
\renewcommand{\algorithmicrequire}{ \textbf{Output}}
\Require Rearranged features $\mathbf{H}_r \in \mathbb{R}^{d \times B}$. 
\State $b_a \gets  \lceil\frac{b}{w}\rceil\times w - b$
\Comment{Padding width}
\State $B \gets  b + b_a$
\Comment{Length after padding}
\State $\mathbf{H}_s \gets \operatorname{Reflect Padding}\left(\mathbf{H}_i, \operatorname{width=}\left(\lfloor\frac{b_a}{2}\rfloor, b_a-\lfloor\frac{b_a}{2}\rfloor\right)\right)$
\State $\mathbf{Z}_s \gets \operatorname{Reflect Padding}\left(\mathbf{Z}_i, \operatorname{width=}\left(\lfloor\frac{b_a}{2}\rfloor, b_a-\lfloor\frac{b_a}{2}\rfloor\right)\right)$
\State $\mathbf{Z}_s \gets {\mathbf{Z}_s}/{256}$ \Comment{Scale the original coordinates}
\State $\mathbf{Z}_s \gets \left[\mathbf{Z}_s - \left(x_{min},y_{min}\right)\right]+1$ 
\Comment{$\left(x,y\right)$ is the coordinate}
\State Initialize $\mathbf{H}_r$ as $\emptyset$
\For{$idx \in \left[0: B: w\right]$}

\State $\triangleright$ Select the $w$ features closest to $\boldsymbol{z}_{s,1}$ in $\mathbf{Z}_s$, including $\boldsymbol{z}_{s,1}$ itself
\State $select\_idx \gets \operatorname{Hnsw.query}\left(\boldsymbol{z}_{s,1},\operatorname{topn=}w\right)$ 
\State $\triangleright$ Add the $w$ closest features to the new array
\State $\mathbf{H}_r \gets \mathbf{H}_r+\mathbf{H}_s\left[select\_idx\right]$ 
\State $\mathbf{H}_s \gets
\mathbf{H}_s - \mathbf{H}_s[select\_idx]$
\Comment{Delete selected $\boldsymbol{h}$}
\State $\mathbf{Z}_s \gets
\mathbf{Z}_s - \mathbf{Z}_s[select\_idx]$ \Comment{Delete selected $\boldsymbol{z}$}
\EndFor
\end{algorithmic}
\end{algorithm}

\subsubsection{Random window masking} 
To increase the robustness of the model for tumor heterogeneity and further exploit the advantages of our hierarchical processing framework, we propose a random window masking strategy. A WSI bag will be further split into several sub-WSI bags.
Inspired by the superpixel sampling strategy  \cite{bian2022multiple}, we sample at the window level to maintain 2D spatial information in each local window. 
Specifically, we perform $m$ random window sampling for the rearranged feature sequence. A WSI feature is divided into $m$ sub-WSIs for subsequent feature aggregation. To avoid adding additional computational burden, the feature number of each sub-WSI is $1/m$ of the original WSI.

\subsection{Feature Aggregation}
To better encode the spatial, contextual and hierarchical information in the patient-level bag, we propose a hierarchical vision Transformer named HVTSurv to perform feature aggregation in the patient-level bag. The HVTSurv is mainly composed of three layers, including the local-level, WSI-level and patient-level interaction layer. In our paper, local-level means patch features within the same window, WSI-level means patch features from different local windows within a sub-WSI, and patient-level means patch features from different sub-WSIs within a patient. 
The overview of proposed three interaction layers is shown in Fig. \ref{figA1}.

\subsubsection{Local-level interaction layer}

To encode local spatial information in each WSI, we design a local-level interaction layer.
Due to the irregularly shaped property of WSIs, the local windows usually appear irregularly shaped. In our intuitive experience, the distance information in the local space always contains more near range spatial structure information than the direction information in WSI. So in this paper, we use the Manhattan distance to encode the relative position information between different patches in each window. The 2D spatial information between different patches is consequently reduced to 1D distance information. Similar to the relative position encoding method used in Swin-Transformer \cite{liu2021Swin}, a learnable matrix $\hat{\mathbf{B}}$ is used to learn the embedding of different distances, which is combined with the self-attention (SA). In the local window block, the self-attention \cite{liu2021Swin} corresponding to each head in computing similarity can be defined as:
\begin{equation}
\operatorname{SA}_{\text{local}}=\operatorname{softmax}\left(\frac{\mathbf{Q}\mathbf{K}^T+\mathbf{B}}{\sqrt{d}}\right),
\label{eqC1}
\end{equation}
where $\mathbf{Q}\in \mathbb{R}^{w \times d}$, $\mathbf{K}\in \mathbb{R}^{w \times d}$, $\mathbf{B}\in \mathbb{R}^{w \times w}$ is the relative position bias, and
values in $\mathbf{B}$ are taken from $\hat{\mathbf{B}}$, with $w$ being the number
of patch features in a window. \par
A segmented Manhattan distance is used to make the model more sensitive to short rather than long distances. Inspired by the method in  \cite{wu2021rethinking}, the expression of the piecewise function is defined as follows:
\begin{equation}
g(x)= \begin{cases}{[|x|],} & |x| \leq \alpha \\
\min \left(\lambda,\left[\alpha+\frac{\ln (|x| / \alpha)}{\ln (\gamma / \alpha)}(\beta-2\alpha)\right]\right), & |x|>\alpha\end{cases}
\label{eqC2}
\end{equation}
where $[\cdot]$ is a round operation, $\alpha$, $\beta$, $\lambda$, $\gamma$ are all hyperparameters and we parameterize a learnable matrix $\hat{\mathbf{B}}\in \mathbb{R}^{(2\lambda+1) \times \text{head}}$ for all heads. \par

\subsubsection{WSI-level interaction layer}
To encode WSI-level long-distance contextual information, we design a WSI-level interaction layer. We adopt the spatial shuffle method so the patch features from different regions in a WSI-level bag can be used for similarity computation in the same window. Specifically, for each WSI-level bag after the local-level interaction layer,
we spatially shuffle the feature sequence before dividing the window and calculating the window attention.
It should be noted that in the self-attention calculation, we do not add spatial information. For the spatial shuffle algorithm, we use the shuffle method noted in  \cite{huang2021shuffle}. In shuffle window block, the self-attention \cite{vaswani2017attention} corresponding to each head in computing similarity can be defined as:
\begin{equation}
\operatorname{SA}_{\text{shuffle}}=\operatorname{softmax}\left(\frac{\mathbf{Q}\mathbf{K}^T}{\sqrt{d}}\right),
\label{eqC3}
\end{equation}
where $\mathbf{Q}\in \mathbb{R}^{w \times d}$, $\mathbf{K}\in \mathbb{R}^{w \times d}$, with $w$ being the number
of patch features in a window. \par 
\subsubsection{Patient-level interaction layer}
To further explore the hierarchical information from the WSI to the patient level,  we design a patient-level interaction layer focusing on global contextual interaction across the entire patient-level bag.
Specifically, we first concatenate all sub-WSI features corresponding to a patient, and then an attention pooling layer (AttnPool) is used to obtain patient-level representation $\boldsymbol{h}_{\text {patient}}$ to estimate the patient's hazard risk $h(k \mid  \mathbf{X}_{i})$. Specifically, AttnPool can be defined as:

\begin{equation}
\begin{aligned}
&a_{g}=\frac{\exp \left\{\mathbf{U}\left(\tanh \left(\mathbf{V}\boldsymbol{h}_{g}\right)\right)\right\}}{\sum_{j=1}^{G} \exp \left\{\mathbf{U}\left(\tanh \left(\mathbf{V} \boldsymbol{h}_{j}\right)\right)\right\}}, \\
&\boldsymbol{h}_{\text{patient}}=\sum_{g=1}^{G} a_{g} \boldsymbol{h}_{g},
\end{aligned}
\end{equation}
where $\mathbf{U}\in \mathbb{R}^{1 \times d_{h}}$, $\mathbf{V}\in \mathbb{R}^{d_{h} \times d}$, with $d_{h}$ being the dimension of hidden layer, $G$ is the number of patches in a patient. \par 

To optimize the model parameters, we adopt the log likelihood function \cite{zadeh2020bias,chen2021pan} as loss function. For an uncensored patient $(C_i=0)$ with failure in interval $[t_k, t_{k+1})$, the likelihood can be calculated as the survival probability in $[t_0, t_{k})$ multiplied by the failure probability in $[t_k, t_{k+1})$:
\begin{equation}
l_{\text{uncensored}} = h(k \mid  \mathbf{X}_{i})S(k-1 \mid  \mathbf{X}_{i}).
\label{eq4}
\end{equation}
For a censored patient $(C_i=1)$ with censored in interval $[t_k, t_{k+1})$, the likelihood can be calculated as the survival probability in $[t_0, t_{k+1})$:
\begin{equation}
l_{\text{censored}} = S(k \mid  \mathbf{X}_{i}).
\label{eq5}
\end{equation}
Finally, the loss function can be defined as:
\begin{equation}
\begin{aligned}
L=&-C_i\operatorname{log}S(k \mid  \mathbf{X}_{i})\\
&-(1-C_i)\operatorname{log}S(k-1 \mid  \mathbf{X}_{i})\\
&-(1-C_i)\operatorname{log}h(k \mid  \mathbf{X}_{i}).\\
\end{aligned}
\label{eq6}
\end{equation}

\section{Experimental Results}
\subsection{Datasets}
We closely follow the data settings of PatchGCN \cite{chen2021whole}. Five public cancer types from TCGA are adopted: Bladder Urothelial Carcinoma (BLCA), Breast Invasive Carcinoma (BRCA),  Glioblastoma\&Lower Grade Glioma (GB\&LG), Lung Adenocarcinoma (LUAD), Uterine Corpus Endometrial Carcinoma (UCEC). We take gastrointestinal tract cancer type into our experiment for a comprehensive comparison: Colon\&Rectal Adenocarcinoma (CO\&RE). 
Six public cancer datasets include 3,104 patients and 3,752 H\&E diagnostic WSIs, whose specific information is summarized in Table~\ref{tab1}.

\begin{table}[]
\centering

\label{tabb1}
\begin{threeparttable}
\setlength{\tabcolsep}{1mm}
\begin{tabular}{c|cccc}
\hline
Cancer Type & Patient & WSI & Censored$^1$ & Time (month)$^2$ \\ \hline
BLCA        & 373        & 437    & 0.547 &  163.2   \\
BRCA        & 956        & 1,022   & 0.864 &  282.7   \\
CO\&RE    & 339        & 344    & 0.764   &  147.9 \\
GB\&LG      & 519        & 895   & 0.690  &  211.0  \\
LUAD        & 452        & 514    & 0.650  &  238.1  \\
UCEC        & 465        & 540    & 0.837  &  225.5  \\ 
\hline
\end{tabular}
\begin{tablenotes}
 \item[1] The ratio of censored patients in the dataset.
 \item[2] The longest survival time of patients in the dataset.
\end{tablenotes}
\end{threeparttable}
\caption{Datasets Summary}
\label{tab1}
\end{table}

\subsection{Evaluation Metric and Implementation Details}
This paper uses Concordance Index (C-Index) and Kaplan–Meier (KM) estimator with a Log-rank test for evaluation metrics. For the dataset partition, we adopt 4-fold cross-validation. For WSIs and follow-up labels, we follow the PatchGCN processing step. For the parameters and training of HVTSurv, the window size is 49, the number of sub-WSIs is 2, and the survival loss function Eq.(\ref{eq6}) is adopted with the training batch size being 1. More details are in Appendix.

\begin{table*}[h]
\centering

\setlength{\tabcolsep}{3.1mm}
\begin{tabular}{c|cccccc|c}
\hline
\textbf{}                                                                 & \textbf{BLCA}                      & \textbf{BRCA}                      & \textbf{CO\&RE}                  & \textbf{GB\&LG}                    & \textbf{LUAD}                      & \textbf{UCEC}                      & \textbf{Mean}  \\ \hline
AMIL[1]                                                    & ${0.499}_{.015}$         & $0.571_{.037}$         & $0.543_{.038}$        & $\underline{0.756}_{.117}^{\dagger}$          & $0.548_{.063}$          & $0.561_{.069}$           & $0.580$         \\
DSMIL[2] & ${0.530}_{.064}$          & $\underline{0.575}_{.048}^{\dagger}$        & $0.571_{.085}$          & ${0.734}_{.133}^{\dagger}$ & ${0.562}_{.048}^{\dagger}$         & $0.612_{.091}^{\dagger}$         & $0.597$         \\
TransMIL[3]                                    & $\underline{0.572}_{.021}$        & $0.548_{.067}$         & $0.588_{.051}^{\dagger}$          & $0.748_{.117}^{\dagger} $         & $0.519_{.057}$          & $0.616_{.051}^{\dagger} $         & $0.599$          \\
ESAT[4]                                    & $0.562_{.027}^{\dagger}$          & ${0.516}_{.035}$  &$ 0.562_{.097}$          & ${0.489}_{.039}$         & $0.533_{.031}$          & $0.463_{.036}$          & $0.521$        \\
DeepAttnMISL[5]                                   & $0.491_{.040}$        & $0.571_{.046}^{\dagger}$      & ${0.536}_{.031}$ & $0.697_{.157}^{\dagger}$         & $0.561_{.045}$         & ${0.576}_{.079}^{\dagger}$         & ${0.572}$        \\
SeTranSurv[6] & ${0.549}_{.019}$ & $0.547_{.051} $        & $0.536_{.038}$         & $0.682_{.138}^{\dagger}$         & $0.560_{.049}^{\dagger} $        & ${0.601}_{.066}^{\dagger}$ & $0.579 $     \\
DeepGraphSurv[7]                                                    & $0.535_{.047} $         & $0.570_{.076}^{\dagger}$       & $0.585_{.056}^{\dagger}$         & $0.737_{.138}^{\dagger}$       & $\underline{0.569}_{.041} $       & $0.580_{.073}^{\dagger}$        & $0.596$        \\
PatchGCN[8]                                                    & $0.544_{.019} $         & $0.568_{.040}^{\dagger}$       & $\underline{0.599}_{.068}^{\dagger}$         & $0.743_{.107}^{\dagger}$       & $0.567_{.081}^{\dagger} $       & $\underline{0.632}_{.030}^{\dagger}$        & $\underline{0.609}$        \\

HVTSurv                                                                  & $\textbf{0.579}_{.019}^{\dagger}$         & $\textbf{0.614}_{.037}^{\dagger}$        & $\textbf{0.606}_{.084}^{\dagger}$         & $\textbf{0.779}_{.019}^{\dagger}$       & $\textbf{0.584}_{.015}^{\dagger}$ & $\textbf{0.643}_{.058}^{\dagger}$          & $\textbf{0.634}$ 
\\ \hline
\end{tabular}
\caption{Comparison of C-Index performance in TCGA. (``$\dagger$" denotes P-Value \textless 0.05) [1] \cite{chen2021pan}, [2] \cite{li2020dual}, [3] \cite{shao2021transmil}, [4] \cite{shen2022explainable}, [5] \cite{yao2020whole}, [6] \cite{huang2021integration}, [7] \cite{li2018graph}, [8] \cite{chen2021whole}}
\label{tabD1}
\end{table*}

\begin{table*}[h]
\centering

\setlength{\tabcolsep}{1.7mm}
\begin{tabular}{c|cccccc|c}
\hline
\textbf{}                                                                 & \textbf{BLCA}                      & \textbf{BRCA}                      & \textbf{CO\&RE}                  & \textbf{GB\&LG}                    & \textbf{LUAD}                      & \textbf{UCEC}                      & \textbf{Mean}  \\ \hline
 w/o position encoding                                                     & $\underline{0.579}_{.012}$         & $\underline{0.603}_{.046}^{\dagger}$         & $0.565_{.076}^{\dagger}$        & $0.767_{.030}^{\dagger}$          & $0.552_{.034}$          & $0.610_{.026}^{\dagger}$           & $0.613$          \\
  w/o spatial shuffle                                                     & $\textbf{0.582}_{.014}^{\dagger}$         & $0.599_{.061}^{\dagger}$         & $\underline{0.599}_{.076}^{\dagger}$        & $0.776_{.030}^{\dagger}$          & $0.564_{.017}^{\dagger}$          & $0.634_{.047}^{\dagger}$           & $\underline{0.626}$          \\ \hline
  w/o local-level interaction layer                                    & $\underline{0.579}_{.011}$        & $0.575_{.070}$         & $0.592_{.098}^{\dagger}$          & $0.765_{.019}^{\dagger} $         & $0.579_{.008}^{\dagger}$          & $0.638_{.040}^{\dagger} $         & $0.621$          \\ 
 w/o WSI-level interaction layer& $0.536_{.079}$          & $0.599_{.046}^{\dagger}$        & $0.578_{.067}$          & $\textbf{0.785}_{.033}^{\dagger}$& $\underline{0.582}_{.009}^{\dagger}$         & $0.624_{.026}^{\dagger}$         & $0.617$          \\
  w/o patient-level interaction layer                                                    & $0.573_{.024}^{\dagger} $         & $0.593_{.049}^{\dagger}$       & $0.591_{.097}^{\dagger}$         & $\underline{0.779}_{.014}^{\dagger}$       & $0.556_{.014}^{\dagger} $       & $\textbf{0.643}_{.037}^{\dagger}$        & $0.622$          \\
 \hline
HVTSurv                                                                   & $\underline{0.579}_{.019}^{\dagger}$         & $\textbf{0.614}_{.037}^{\dagger}$        & $\textbf{0.606}_{.084}^{\dagger}$         & $\underline{0.779}_{.019}^{\dagger}$       & $\textbf{0.584}_{.015}^{\dagger}$ & $\textbf{0.643}_{.058}^{\dagger}$          & $\textbf{0.634}$
\\ \hline
\end{tabular}
\caption{Effect of different modules and major components in HVTSurv. (``$\dagger$" denotes P-Value \textless 0.05)}
\label{tabD2}
\end{table*}

\begin{table}[h]
\centering
\begin{threeparttable}

\setlength{\tabcolsep}{0.3mm}
\begin{tabular}{c|cccccc|c}
\hline
& \textbf{BLCA}               & \textbf{BRCA}             & \textbf{CO\&RE}         & \textbf{GB\&LG}           & \textbf{LUAD}             & \textbf{UCEC}             & \textbf{Mean}              \\\hline
25  & $\underline{0.585}$ & $0.578^{\dagger}$ & ${0.601}^{\dagger}$ & ${0.769}^{\dagger}$ & $\textbf{0.588}^{\dagger}$ & $0.638^{\dagger}$ & $\underline{0.626}$         \\
36 & ${0.576}^{\dagger}$ & $0.595^{\dagger}$ & $\underline{0.604}^{\dagger}$ & $0.760^{\dagger}$ & ${0.572}^{\dagger}$ & $\underline{0.642}^{\dagger}$  & ${0.625}$         \\
64 & $0.581^{\dagger}$ & $\underline{0.603}^{\dagger}$ & $0.571$ & $\underline{0.777}^{\dagger}$ & $0.574^{\dagger}$ & $0.635^{\dagger}$ & $0.624$         \\
81  & $\textbf{0.588}^{\dagger}$ & $0.598^{\dagger}$ & ${0.569}^{\dagger}$ & $0.774^{\dagger}$ & $0.553^{\dagger}$ & ${0.627}^{\dagger}$ & ${0.618}$         \\
49$^{1}$           & $0.579^{\dagger}$ & $\textbf{0.614}^{\dagger}$ & $\textbf{0.606}^{\dagger}$ & $\textbf{0.779}^{\dagger}$ & $\underline{0.584}^{\dagger}$ & $\textbf{0.643}^{\dagger}$ & $\textbf{0.634}$  
\\ \hline
\end{tabular}
   \begin{tablenotes}
     \item[1] In our paper, we use a window size of 49.
   \end{tablenotes}
\end{threeparttable}
\caption{Effect of different window size in HVTSurv. (``$\dagger$" denotes P-Value \textless 0.05)}
\label{tabD3}
\end{table}

\begin{table}[h]
\centering
\begin{threeparttable}

\setlength{\tabcolsep}{0.4mm}
\begin{tabular}{c|cccccc|c}
\hline
& \textbf{BLCA}               & \textbf{BRCA}             & \textbf{CO\&RE}         & \textbf{GB\&LG}           & \textbf{LUAD}             & \textbf{UCEC}             & \textbf{Mean}              \\\hline
1  & $\underline{0.584}^{\dagger}$ & $0.607^{\dagger}$ & $0.581$ & $\underline{0.778}^{\dagger}$ & $0.575^{\dagger}$ & $0.631^{\dagger}$ & $0.626$         \\
3 & $\textbf{0.585}^{\dagger}$ & ${0.601}^{\dagger}$ & ${0.586}$ & $0.773^{\dagger}$ & $\textbf{0.586}^{\dagger}$ & $0.638^{\dagger}$ & $0.628$         \\
4 & ${0.582}^{\dagger}$ & $\textbf{0.618}^{\dagger}$ & $0.590$ & $0.774^{\dagger}$ & $0.567^{\dagger}$ & $0.633^{\dagger}$ & $0.627$         \\
5  & $0.580$ & $0.613^{\dagger}$ & $\underline{0.597}$ & $0.775^{\dagger}$ & ${0.580}^{\dagger}$ & $\underline{0.639}^{\dagger}$ & $\underline{0.631}$         \\
2$^{1}$           & $0.579^{\dagger}$ & $\underline{0.614}^{\dagger}$ & $\textbf{0.606}^{\dagger}$ & $\textbf{0.779}^{\dagger}$ & $\underline{0.584}^{\dagger}$ & $\textbf{0.643}^{\dagger}$ & $\textbf{0.634}$  
\\ \hline
\end{tabular}
   \begin{tablenotes}
    \item[1] In our paper, we sample 2 sub-WSIs for each WSI feature. The masking ratio for each WSI feature is 0.5. 
   \end{tablenotes}
\end{threeparttable}
\caption{Effect of different sub-WSI numbers for the random window masking strategy.  (``$\dagger$" denotes P-Value \textless 0.05)}
\label{tabD4}
\end{table}

\subsection{Results and Discussion}
Performance comparisons for all methods are summarized in Table \ref{tabD1} and Appendix Fig. 3, including C-Index scores, P-Values and KM analysis. We also compare computational efficiency in Appendix Table 1. For the 4-fold cross-validation C-Index results in Table \ref{tabD1}, we present it as ``$\text{average C-Index}_{\text{standard deviation}}$''. ``TCGA-Mean'' represents the average C-Index scores on the 6 TCGA datasets. 
Besides, we bold the \textbf{best} and underline the \underline{second best}.\par
Compared with WSI-level MIL methods such as DSMIL and TransMIL, the results of patient-level MIL methods including PatchGCN and HVTSurv show that hierarchically aggregating the patient-level features can make a better survival prediction.
Compared with random sampling methods such as DeepAttnMISL and SeTranSurv, HVTSurv adopts the hierarchical processing framework that can handle more patch features. Moreover, local spatially correlated windows obtained by the feature rearrangement can help to achieve significantly better results.
Compared with Transformer-based methods such as TransMIL, SeTranSurv, and ESAT, convolutional based and sin-cos based position encoding schemes pay more attention to global spatial information, which inevitably loses local prognostic information. HVTSurv creatively adopts the Manhattan distance to represent the relative position in the local window, 
which can correctly and effectively encode local prognostic information.
Compared with GNN-based models such as DeepGraphSurv and PatchGCN, different from simply increasing model depth, HVTSurv adopts spatial shuffle for all the local windows, which can encode WSI-level interaction more efficiently.
Compared with other methods in computational efficiency, HVTSurv benefits from the window attention method and has more efficient GPU Memory Costs in patient-level MIL task.
Compared with other methods in Log-rank test, binary experiments show that low and high-risk patients have a statistically signiﬁcant difference (P-Value $<$ 0.05) over 6 cancer types.
In summary, the average C-Index is 2.50-11.30\% higher than all competitive models over 6 TCGA datasets.

\subsection{Ablation and Effectiveness Analysis}

We further conduct a series of ablation studies to determine the contribution of different modules and major components in HVTSurv and test the parameters used in this paper. 
We use the average C-Index score to measure the performance.\par

\begin{figure*}[h]
\centering
\includegraphics[width=1.0\linewidth]{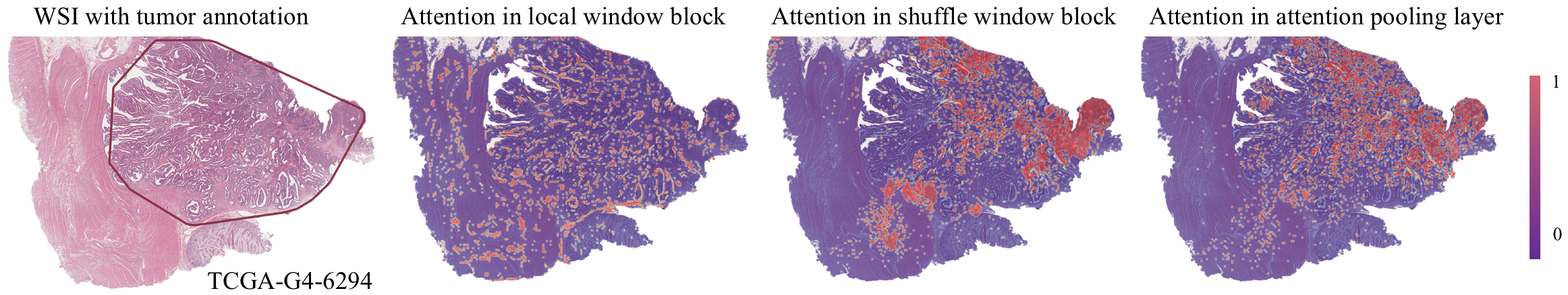}
\caption{Visual analysis of attention in three interaction layers. 
The doctor-annotated cancer areas are shown in dark red.}
\label{figD2_slide}
\end{figure*}
 
\begin{figure}[h]
\centering
\includegraphics[width=1.0\linewidth]{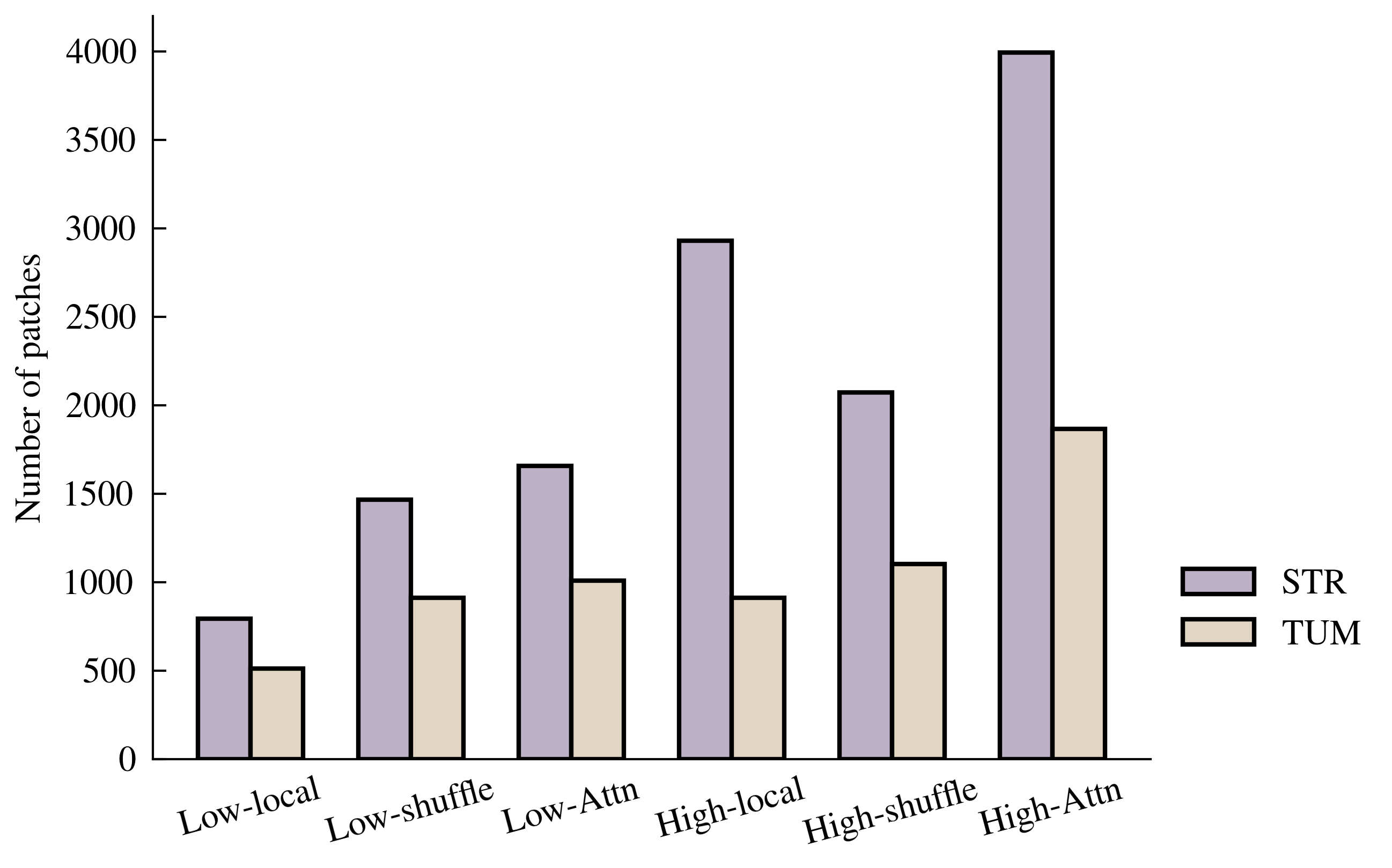}
\caption{Tissue classification results of highly concerned patches in CO\&RE. For clarity, we show the number of patches for tumor-associated tissues including cancer-associated stroma (STR) and colorectal adenocarcinoma epithelium (TUM). ``Low/High-local/shuffle/Attn" represents the number of patches for three interaction layers in low/high-risk patients, respectively.}
\label{figD2_patch}
\end{figure}

In Table \ref{tabD2} and Appendix Table 2,
we test the effect of different modules, major components and different model structures in HVTSurv.
The results show that both position encoding and spatial shuffle play a significant role in improving the performance of the model. 
WSI is a high-resolution and irregularly shaped image after background removal, it is hard to directly encode accurate contextual interaction in the WSI-level bag. So we devise a two-step approach, i.e., local position encoding and WSI-level spatial shuffle. Local accurate spatial information is an essential basis for global information encoding, we find it has a more significant effect on the performance improvement.
Besides, we also perform ablation experiments on the three major components. In most cancers, combining the three interaction layers can better encode the spatial, contextual and hierarchical information in the patient-level bag. Due to the different prognostic features of various cancers, the importance of each interaction layer is slightly different. BLCA and UCEC are two similar cancer types whose prognosis depend more on global-level features such as the depth of tumor invasion in the myometrium and bladder wall, so the WSI-level interaction layer plays a more critical role in these cancer types.
In Table \ref{tabD3}, 
we test the effect of different window sizes in HVTSurv,
we find that a moderate window size can help the model to learn the spatial interaction within the window more accurately and efficiently. Moreover, a relatively small window size can fully utilize the window method's high computational efficiency.
In Table \ref{tabD4}, 
we test the effect of different sub-WSI numbers for the random window masking strategy, 
it can be found that the training strategy of random window masking can help the model better adapt to the heterogeneity of cancer. Moreover, dividing a WSI into multiple sub-WSIs 
can further exploit the advantages of our hierarchical processing framework.

\subsection{Interpretability and Attention Visualization}

We further explore the interpretability of our HVTSurv model in the slide level and patch level, whose results are shown in Fig. \ref{figD2_slide} and Fig. \ref{figD2_patch}, respectively. The details are in the Appendix. In Fig. \ref{figD2_slide}, the attention to the three major components is gradually extended from local low-level information to global high-level information, such as cancer-related regions.
Benefiting from the hierarchical network, the receptive field of the model can gradually become larger, then the hierarchical information in the patient-level bag can be fully explored.
Besides, in Fig. \ref{figD2_patch}, we show the number of tumor-related patches in the high attention score area, which further explains from patch level statistical result for the whole CO\&RE dataset. HVTSurv adopts a hierarchically designed network structure encoding interactions from local-level to WSI-level and further to patient-level, which can gradually discover the critical prognostic tissues like tumor-related tissues STR and TUM. We can also find that for high-risk patients, 
the number of tumor-related patches has significantly increased, which has been medically proven to be related to the prognosis of colorectal cancer \cite{abbet2020divide}.

\section{Conclusion}
In  this  work,  we  propose  a  hierarchical Vision Transformer named HVTSurv  that  progressively  explores  local-level spatial interaction, WSI-level contextual interaction and patient-level hierarchical interaction in the  patient-level survival prediction. 
Hierarchical processing framework effectively reduces the computational cost, which is suitable for patient-level MIL tasks. Inspired by this, we propose the local and shuffle window block to progressively obtain the WSI-level representation and an attention pooling layer to get patient-level hazard risk. Besides, we design feature pre-processing strategies, including feature rearrangement and random window masking to explore spatial, contextual, hierarchical information better.
Compared to SOTA methods, we achieve a better average C-Index over the 6 TCGA datasets, with a performance gain of 2.50\%. In  KM  analysis and  Log-rank test, the low and high-risk patients have a statistically significant difference over 6 TCGA cancer types.
In the ablation study, we prove that adopting Manhattan distance based position encoding and spatial shuffle based long-range interaction can obtain  better representation. 
Moreover, the ablation results of three interaction layers demonstrate the effectiveness of our hierarchical processing framework.
The visualization of attention further confirms our conclusions.\par

\section{Acknowledgements}
This work was supported in part by the National Natural Science Foundation of China (61922048\&62031023), in part by the Shenzhen Science and Technology Project (JCYJ20200109142808034), and in part by Guangdong Special Support (2019TX05X187).

\bibliography{references}

\clearpage
\appendix

\section{Method}
\subsection{Feature Generation}
\begin{figure}[H]
\centering
\includegraphics[width=1.0\linewidth]{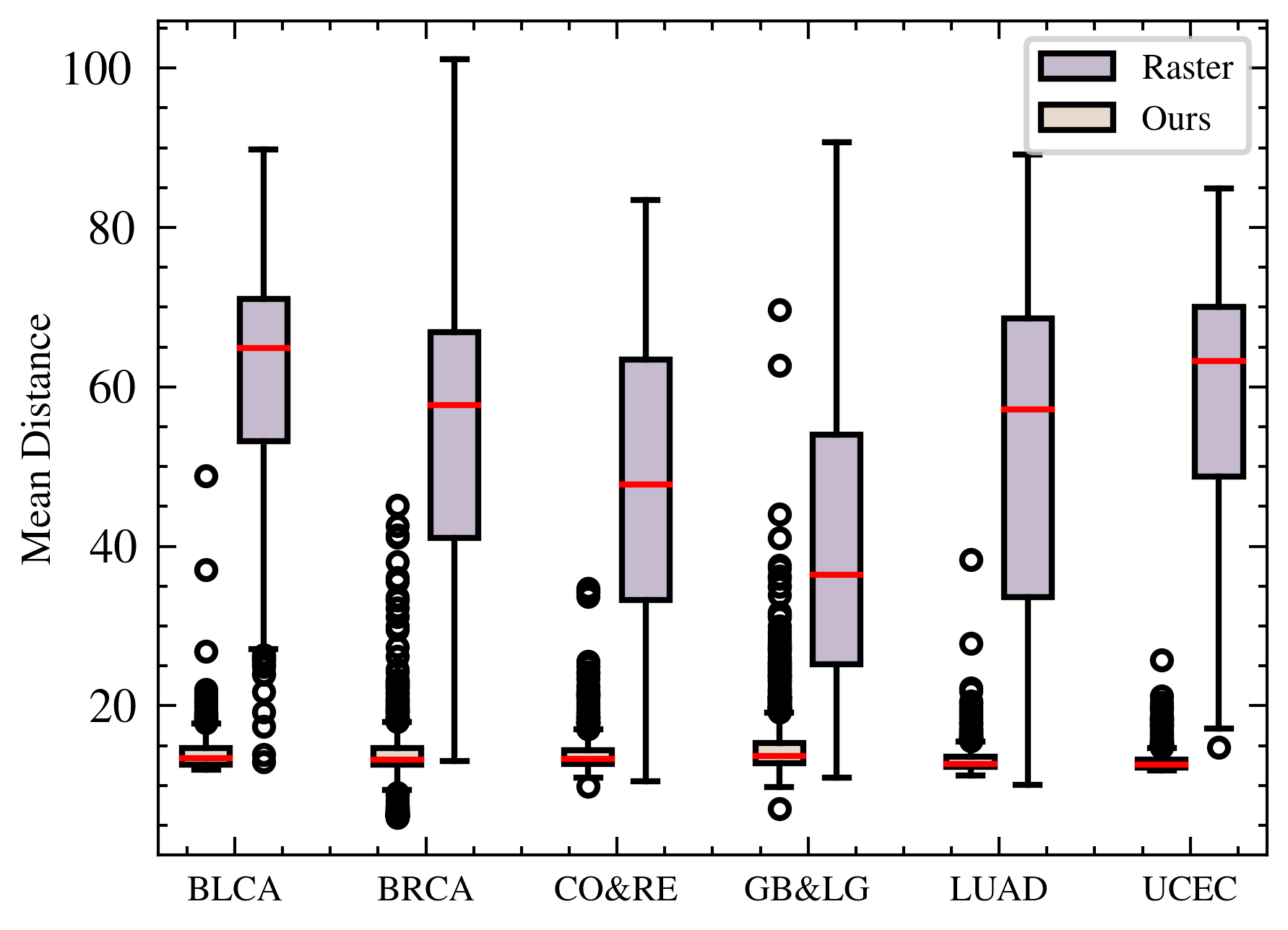}
\caption{Mean distance comparison of different feature expansion methods. We average the sum of Manhattan distances for each window in a WSI as the mean distance. We show the mean distance distribution of different WSIs within each cancer. Boxes represent the data distribution's 1th, median, and 3rd quartiles, with whiskers extending to data points within 1.5 times the interquartile range, and black circles represent outliers.}
\label{figB2}
\end{figure}

\begin{figure*}[h]
\centering
\subfigure[Raw WSI (TCGA-95-A4VK)]{
\begin{minipage}[t]{0.3\linewidth}
\centering
\includegraphics[width=5.3cm]{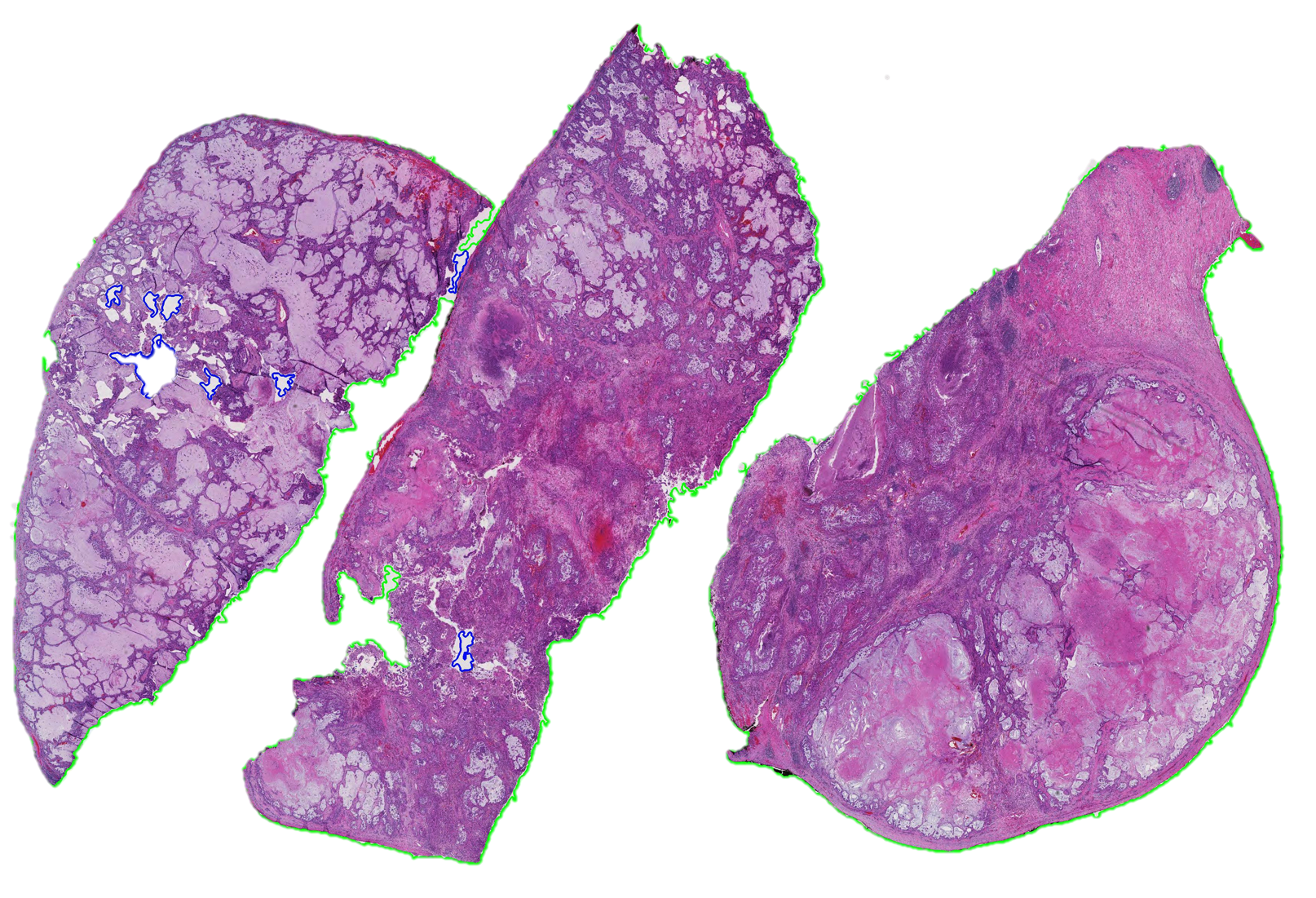}
\end{minipage}%
}%
\subfigure[Raster expansion]{
\begin{minipage}[t]{0.3\linewidth}
\centering
\scalebox{-1}[1]{\includegraphics[width=5.3cm]{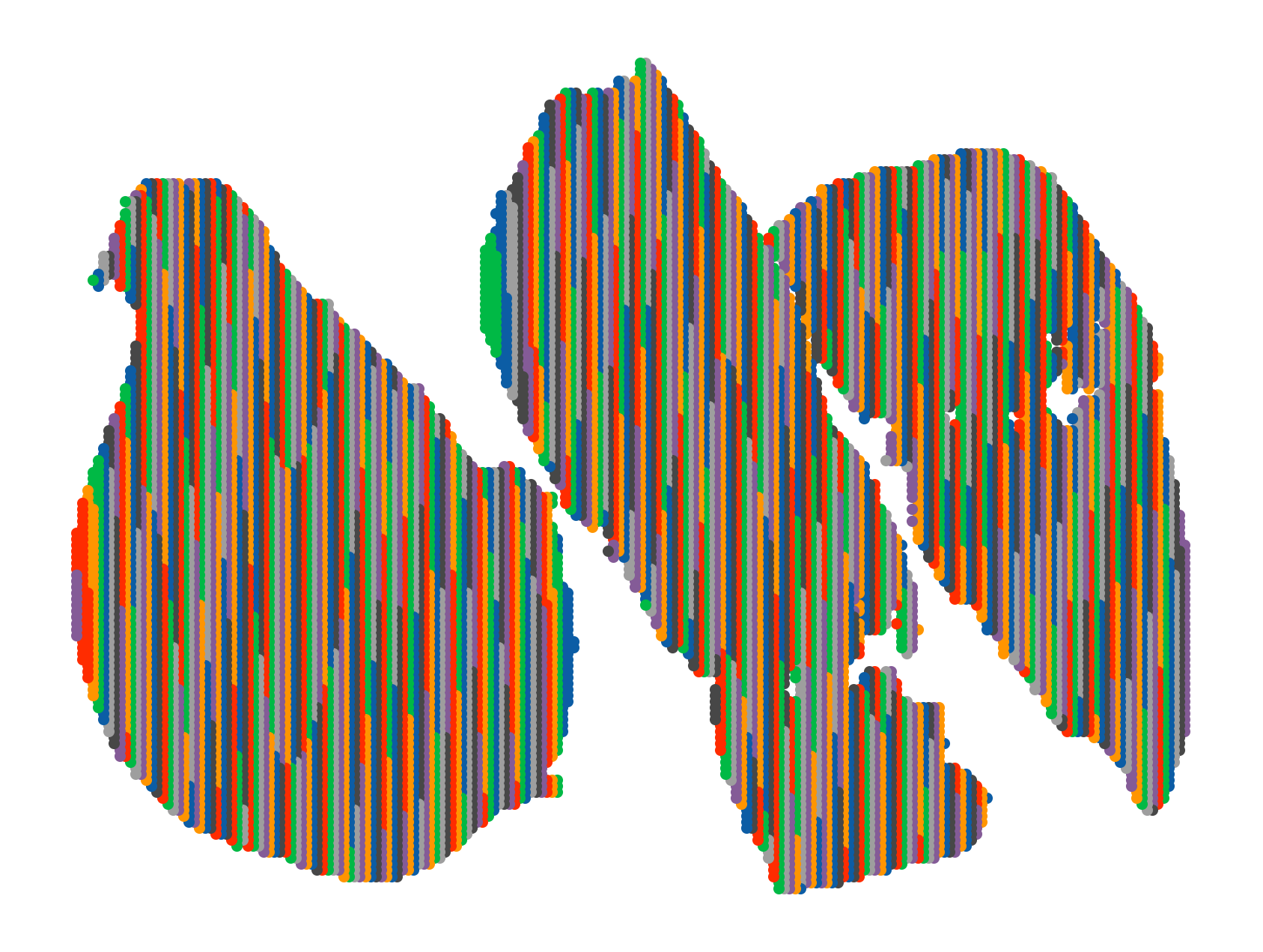}}
\end{minipage}%
}%
\subfigure[Window expansion]{
\begin{minipage}[t]{0.3\linewidth}
\centering
\scalebox{-1}[1]{\includegraphics[width=5.3cm]{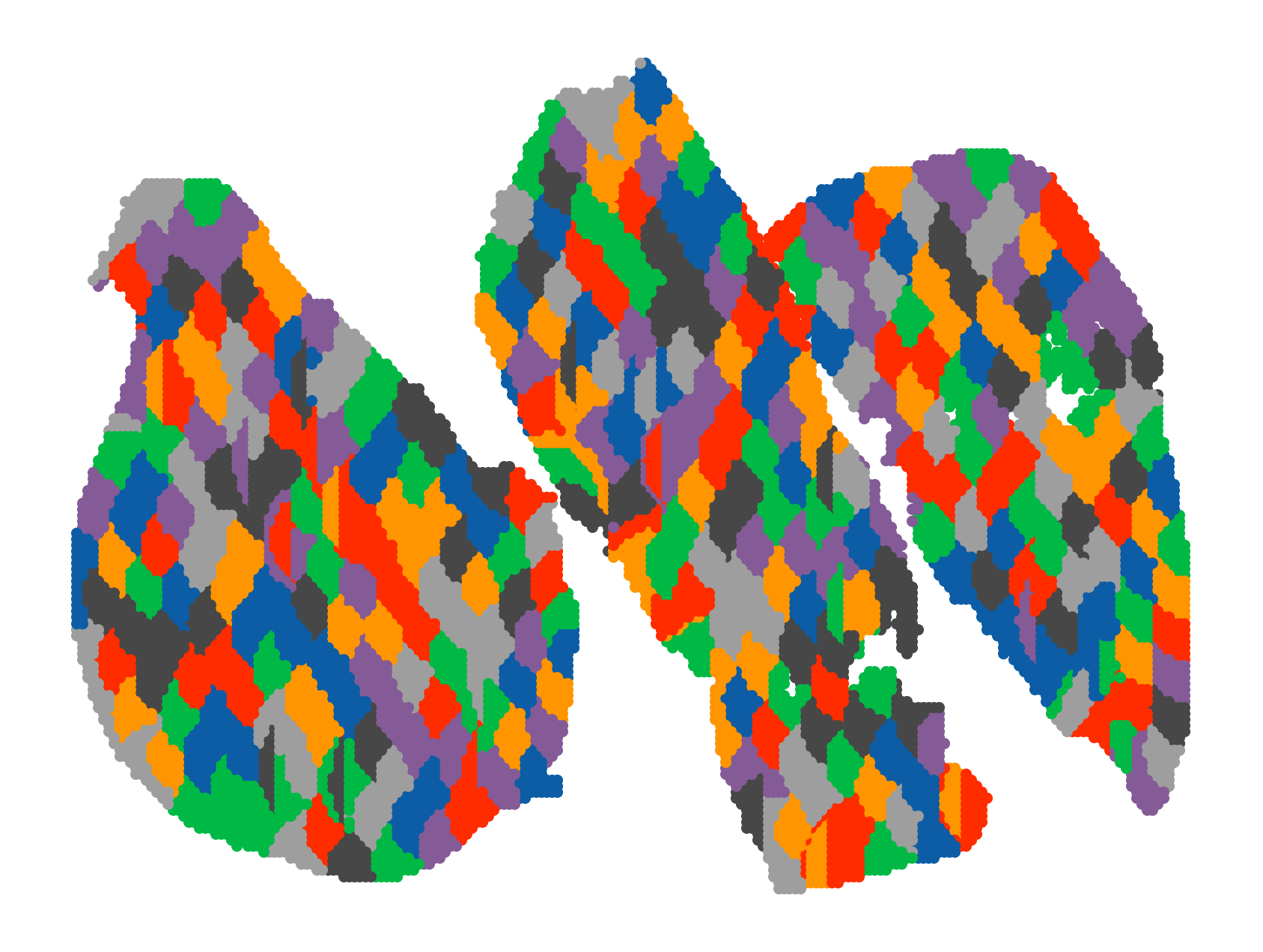}}
\end{minipage}
}%
\caption{Visual comparison of different feature expansion methods. (a) represents the raw WSI, (b) and (c) represent the 2D visualization map of the patch features according to the patch coordinates. Each point in (b) and (c) represents the feature vector of the corresponding patch in (a). Different colors represent different windows, where 7 colors are used alternatively and cycled when visualizing.}
\label{figB1}
\end{figure*}

To better reflect the local characteristics in both horizontal and vertical directions within a window, a feature rearrangement method is proposed to ensure the closeness in both directions of the 2D space after the window partition.
We present quantitative and  qualitative experimental results in Fig. \ref{figB2} and Fig. \ref{figB1}, respectively.

\section{Experimental Results}

\subsection{Evaluation Metric and Statistical Analysis}
\begin{itemize}
\item[$\bullet$]\emph{Concordance Index}
For evaluation metric, Concordance Index(C-Index) \cite{heagerty2005survival} is usually adopted in survival prediction. The C-Index is computed as follows:

\begin{equation}
c=\frac{1}{n} \sum_{i: C_{i}=0} \sum_{j:T_{j}>T_{i}} I\left[f_{i}>f_{j}\right],
\label{eqD1}
\end{equation}

where $n$ is the number of comparable pairs, $I[.]$ is the indicator function, and $f.$ denotes the patient's predicted risk. The value of C-Index ranges from 0 to 1. The larger C-Index value means the model has a better prediction performance. This paper uses the average C-Index to represent cross-validated C-Index performance.

\end{itemize}

\begin{itemize}
\item[$\bullet$]\emph{Kaplan–Meier estimator} For statistical analysis, Kaplan–Meier (KM) estimator \cite{kaplan1958nonparametric} with Log-rank test is a common tool in survival prediction. KM estimator is a non-parametric statistic used to estimate the survival function $S(t)$, which is defined as:
\begin{equation}
\widehat{S}(t)=\prod_{i: t_{i} \leq t}\left(1-\frac{d_{i}}{n_{i}}\right),
\label{eqD2}
\end{equation}
where $t_{i}$ denotes the time when at least one event happened, $d_{i}$ represents the number of events (e.g., deaths) that happened at time $t_{i}$, and $n_{i}$ denotes the individuals known to have survived (have not yet had an event or been censored) up to time $t_{i}$. The Log-rank test is used to test for statistical significance (P-Value $<$ 0.05) in survival distributions between low and high risk patients \cite{bland2004logrank}. We divide high and low risk patients for each test fold according to out-of-sample risk predictions. Then we concatenate all test fold binary results to perform KM analysis and Log-rank test.
\end{itemize}
\subsection{Implementation Details}
For the dataset partition, we use 4-fold cross-validation to train and test all the models for each TCGA cancer type. We randomly split the patient data in the ratio of training: validation: test = 60: 15: 25. We use the StratifiedKFold \cite{scikit-learn} method to ensure the training validation and test sets have similar censorship ratios.
For each WSI, we follow CLAM \cite{lu2021data} processing step to extract patch features, and each WSI contains approximately 12017 256$\times$256 image patches at 20$\times$ magnification, with some patients having up to 17 WSIs. For the random masking strategy, the number of sub-WSI is 2.
For the follow-up label, we follow Patch-GCN \cite{chen2021whole} processing step. For each cancer type, we use the quartiles of survival time for uncensored patients to divide the survival times of all patients into 4 non-overlapping intervals.
For the parameters of HVTSurv, we set $\alpha=1.9, \beta=7.6, \gamma=11.4, \lambda=7$ in the piecewise function. We adopt the feature reduction for the Linear layer from  1024 to 512  dimensions. For each window, the window size is 49, i.e., there are 49 patch features within each window. For the sake of simplicity, the local window block and shuffle window block adopt the same window size. 
Besides,
we instantiate one local-level interaction layer to process all sub-WSIs of a patient separately. 
For the training of HVTSurv, the Ranger optimizer \cite{Ranger} is employed with a learning rate of 2e-4 and weight decay of 1e-5. The validation loss is used as the monitor metric, and the early stopping strategy is adopted, with the patience of 8. In addition, for a fair comparison, we employ the same survival loss function, i.e., cross entropy-based 
survival loss function as Eq.(6) in the main text, ResNet-50 feature embeddings, and training hyperparameters for all the compared methods.

\subsection{Results and Discussion}

To further illustrate the computational efficiency of HVTSurv, we compare FLOPs, Params, and GPU Memory Costs. The TCGA-12-0769 WSI image is selected to calculate FLOPs,
and GB\&LG data set is selected to calculate the GPU Memory Costs under half precision floating-point format, where each WSI contains 12421 patches on average. All experiments are performed on a GeForce RTX 3090, and the results are shown in Table \ref{tabD1_add}. It should be noted that ``XX / 24268 M'' in GPU Memory Cost means the  GPU memory usage during model training, ``24268 M'' represents the total GPU memory of a GeForce RTX 3090. OOM means out of memory, (*) means the input feature dimension is reduced from 1024 to 128, and for others, the input feature dimension is reduced from 1024 to 512.

It can be seen that ESAT and SeTranSurv have many model parameters, and different operations are required to prevent out of memory during model training. Besides, for TransMIL and PatchGCN, it is still challenging to process high-dimensional data. Therefore, for the input features, the dimension is reduced from 1024 to 128 for subsequent feature aggregation. In particular, when the models of TransMIL and PatchGCN have no extra operations, and out of memory will inevitably happen due to large model complexity. Based on the window attention method, HVTSurv is devised to solve the GPU Memory Costs problem in patient-level survival prediction. Under the same feature dimension, HVTSurv has fewer FLOPs, Params, and GPU Memory Costs than PatchGCN. 

\begin{table}[h]
\centering
\caption{Analysis of Computational Efficiency of Different Models}
\label{tabD1_add}
\setlength{\tabcolsep}{0.8mm}
\begin{tabular}{cccc}
\hline
Models        & \multicolumn{1}{c}{FLOPs}  & \multicolumn{1}{c}{Params}  & \multicolumn{1}{c}{GPU Memory Costs} \\ \hline
AMIL          & 3.55 G                     & 0.66 M                      & 3637 / 24268 M                      \\
DSMIL         & 8.54 G                     & 1.59 M                      & 6353 / 24268 M                                 \\
TransMIL   & 15.08 G                    & 2.67 M                      & OOM                                 \\
ESAT          & \multicolumn{1}{c}{6988 G} & \multicolumn{1}{c}{26.25 M} & 11019 / 24268 M                     \\
SeTranSurv    & 5.37 G                     & 8.92 M                      & 24243 / 24268 M                     \\
DeepAttnMISL  & 2.84 G                     & 0.79 M                      & 3433 / 24268 M                      \\
DeepGraphSurv & 12.77 G                    & 2.63 M                      & 12099 / 24268 M                               \\
PatchGCN   & 70.94 G                    & 17.32 M                     & OOM                                 \\
HVTSurv       & 17.82 G                    & 3.28 M                      & 11699 / 24268 M                  \\  \hline
TransMIL(*)      & 1.54 G                     & 0.27 M                      & 9171 / 24268 M   
        \\
PatchGCN(*)      & 4.97 G                     & 1.18 M                      & 19387 / 24268 M   
\\
HVTSurv(*)      & 1.78 G                     & 0.33 M                      & 5015 / 24268 M 
\\ \hline
\end{tabular}
\end{table}

\begin{figure*}[h]
\centering
\includegraphics[width=1.0\linewidth]{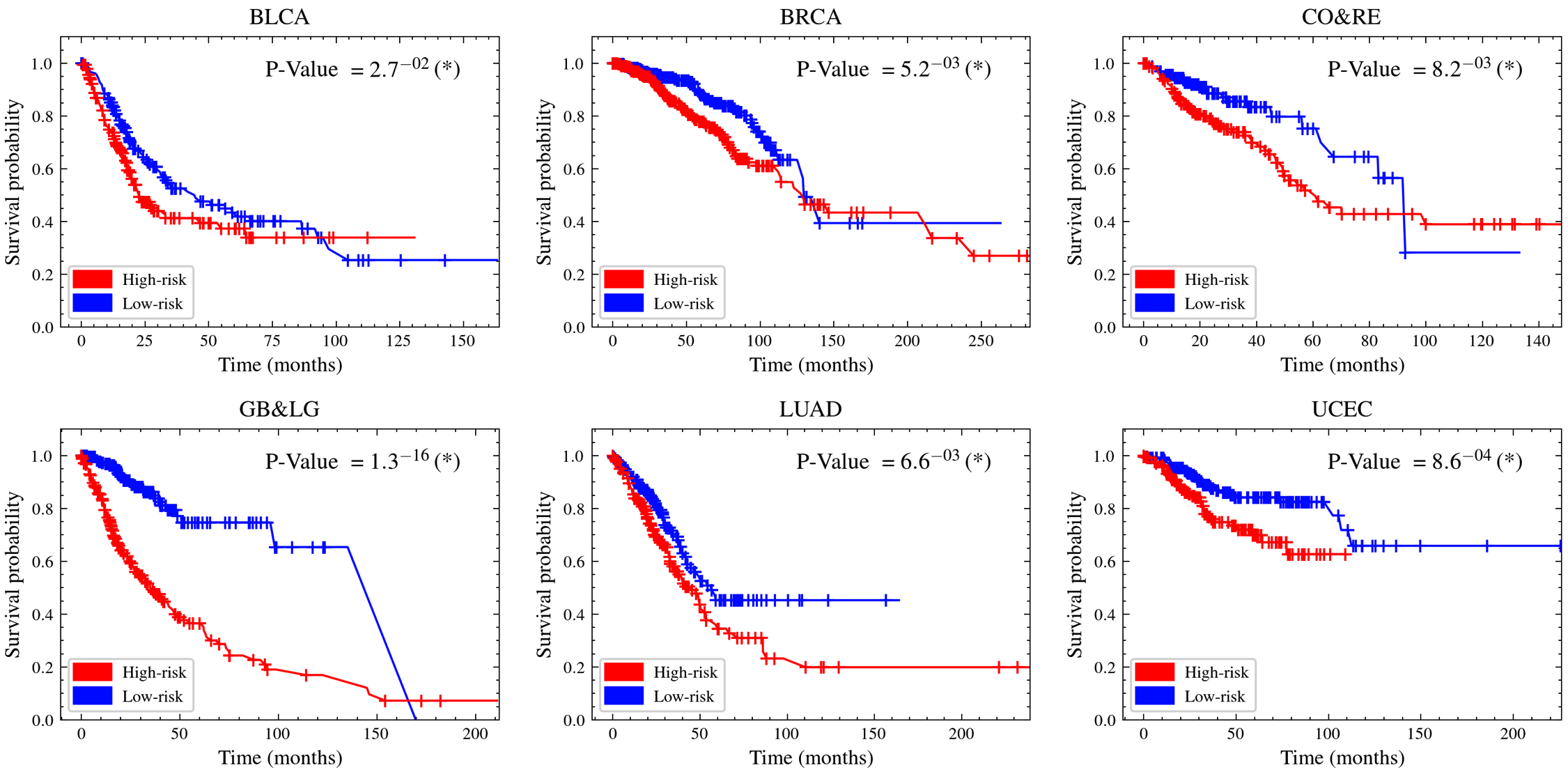}
\caption{KM analysis of HVTSurv. For each test fold, we divide high and low risk patients according to out-of-sample risk predictions. Then we concatenate all the test fold binary results to perform KM analysis. For significance testing, we use the Log-rank test to measure the statistical difference (with $\ast$ marked if P-Value $<$ 0.05) between the two survival distributions corresponding to low-risk (blue) and high-risk (red) patients.}
\label{figA1}
\end{figure*}

\subsection{Ablation and Effectiveness Analyses}
We show the effect of different model structures of HVTSurv in Table \ref{tab2}. 
The results show that the model structures of two WSI-level interaction layers and our HVTSurv model structure can achieve relatively good results.
Since our method employs a two-step progressive encoding of global information, i.e. local accurate spatial information plus global contextual information interaction, the proposed network architecture can achieve more robustness on different cancer datasets compared to other schemes.

\begin{table*}[h]
\centering
\caption{Effect of different model structures in HVTSurv. (``$\dagger$" denotes P-Value \textless 0.05)}
\label{tab2}
\setlength{\tabcolsep}{1.5mm}
\begin{tabular}{cccccccc}
\hline
\textbf{}                                                                 & \textbf{BLCA}                      & \textbf{BRCA}                      & \textbf{CO\&RE}                  & \textbf{GB\&LG}                    & \textbf{LUAD}                      & \textbf{UCEC}                      & \textbf{Mean}  \\ \hline
  local-level interaction layer $\times$2                                     & $0.540_{.010}$          & $\textbf{0.621}_{.054}^{\dagger}$  &$ 0.585_{.043}$          & $\textbf{0.782}_{.023}^{\dagger}$         & $0.572_{.029}^{\dagger}$          & $0.604_{.033}^{\dagger}$          & $0.617$          \\
  WSI-level interaction layer $\times$2                                    & $0.571_{.009}$        & $0.602_{.046}^{\dagger}$      & $\textbf{0.618}_{.084}^{\dagger}$ & $0.763_{.024}^{\dagger}$         & $\underline{0.577}_{.020}^{\dagger}$         & $\underline{0.646}_{.031}^{\dagger}$         & $\underline{0.630}$          \\
interchange order of two interaction layers & $\textbf{0.583}_{.011}$ & $0.596_{.026}^{\dagger} $        & $0.571_{.083}^{\dagger}$         & $0.770_{.022}^{\dagger}$         & $0.572_{.020}^{\dagger} $        & $\textbf{0.649}_{.029}^{\dagger}$ & $0.624 $         \\
HVTSurv                                                                   & $\underline{0.579}_{.019}^{\dagger}$         & $\underline{0.614}_{.037}^{\dagger}$        & $\underline{0.606}_{.084}^{\dagger}$         & $\underline{0.779}_{.019}^{\dagger}$       & $\textbf{0.584}_{.015}^{\dagger}$ & $0.643_{.058}^{\dagger}$          & $\textbf{0.634}$
\\ \hline
\end{tabular}
\end{table*}

\subsection{Interpretability and Attention Visualization}
We further explore the interpretability of our HVTSurv model. For the slide level interpretability, we adopt the TCGA cancer annotation from Loeffler et al \cite{loeffler_chiara_2021_5320076}. For the patch level interpretability, we adopt the nine tissue-class fully supervised dataset NCT-CRC-HE \cite{kather2019predicting} on colorectal cancer for tissue classification of patches with high attention scores in high-risk patients and low-risk patients, respectively. Specifically, for the slide level interpretability, attention weights are re-scaled from min-max to [0, 1], and for better visualization, we drop 80\% of the smaller attention value. For the patch level interpretability, all low-risk and high-risk patients were defined as those below and above the 25\% and 75\% predicted risk percentiles, respectively. For the WSI of low-risk and high-risk patients, 1\% of high-interest patches were selected.

We briefly introduce the visualization. 
For the multi-head self-attention map $A_{multihead} \in \mathbb{R}^{w \times w \times head}$, where $w$ denotes window size, we first utilize a mean pooling in head dimension to obtain $A_{head} \in \mathbb{R}^{w \times w}$. Then we employ a mean pooling in one $w$ dimension to get the importance of each patch feature in the window, i.e., $A \in \mathbb{R}^{1 \times w}$. For better visualization, we first drop 80\% of the smaller attention value, and then splice the results of all the window attention. It  should be noted that for the shuffle window block, the window attention obtained after shuffle needs to be restored to the original spatial position and then spliced. The visualization part also refers to the implementation in CLAM.\par

\end{document}